\documentclass{article}

\usepackage{PRIMEarxiv}

\usepackage[utf8]{inputenc} 
\usepackage[T1]{fontenc}    
\usepackage{hyperref}       
\usepackage{url}            
\usepackage{booktabs}       
\usepackage{amsmath,amsfonts} 
\usepackage{nicefrac}       
\usepackage{microtype}      
\usepackage{fancyhdr}       
\usepackage{graphicx}       
\usepackage{float}          
\usepackage[numbers]{natbib}         
\graphicspath{{media/}}     

 \usepackage{multirow}

\title{Mitigating Instance-Dependent Label Noise: Integrating Self-Supervised Pretraining with Pseudo-Label Refinement
\thanks{\textit{\underline{Citation}}: 
\textbf{Gouranga Bala, Anuj Gupta, Subrat Kumar Behera, Amit Sethi. Mitigating Instance-dependent Label Noise: A Hybrid Method Leveraging Self-supervised Pre-training and Pseudo-label Refinement. Pages.... DOI:000000/11111.}} 
}

\author{
  Gouranga Bala \thanks{Equal contribution}\\
  Department of Electrical Engineering \\
  IIT Bombay \\
  Mumbai, India \\
  \texttt{gouranga.bala23@gmail.com} \\
  \And
  Anuj Gupta \footnotemark[2]\\
  Department of Electrical Engineering \\
  IIT Bombay \\
  Mumbai, India \\
  \texttt{guptaanuj166@gmail.com} \\
  \And
  Subrat Kumar Behera \\
  Department of Electrical Engineering \\
  IIT Bombay \\
  Mumbai, India \\
  \texttt{subrat.725@gmail.com} \\
  \And
  Amit Sethi \\
  Department of Electrical Engineering \\
  IIT Bombay \\
  Mumbai, India \\
  \texttt{asethi@iitb.ac.in} \\
}

\begin{document}
\maketitle

\begin{abstract}
Deep learning models rely heavily on large volumes of labeled data to achieve high performance. However, real-world datasets often contain noisy labels due to human error, ambiguity, or resource constraints during the annotation process. Instance-dependent label noise (IDN), where the probability of a label being corrupted depends on the input features, poses a significant challenge because it is more prevalent and harder to address than instance-independent noise. In this paper, we propose a novel hybrid framework that combines self-supervised learning using SimCLR with iterative pseudo-label refinement to mitigate the effects of IDN. The self-supervised pre-training phase enables the model to learn robust feature representations without relying on potentially noisy labels, establishing a noise-agnostic foundation. Subsequently, we employ an iterative training process with pseudo-label refinement, where confidently predicted samples are identified through a multi-stage approach and their labels are updated to improve label quality progressively. We evaluate our method on the CIFAR-10 and CIFAR-100 datasets augmented with synthetic instance-dependent noise at varying noise levels. Experimental results demonstrate that our approach significantly outperforms several state-of-the-art methods, particularly under high noise conditions, achieving notable improvements in classification accuracy and robustness. Our findings suggest that integrating self-supervised learning with iterative pseudo-label refinement offers an effective strategy for training deep neural networks on noisy datasets afflicted by instance-dependent label noise.
\end{abstract}

\keywords{Instance-dependent Noise \and Self-supervised Learning \and Pseudo-label Refinement}

\section{Introduction}

Deep learning has achieved great success in classification tasks, but it heavily relies on the quality of labeled data. In real-world scenarios, obtaining perfectly labeled data is challenging due to human errors, ambiguities, or resource limitations, often resulting in noisy labels. These noisy labels can significantly degrade model performance, affecting generalization and reliability. Thus, addressing label noise is crucial for building robust models.

Many existing approaches to learning with noisy labels include modifying loss functions, selecting reliable samples, or using multiple models to reduce noise effects. While these techniques perform well under simple, instance-independent noise, they struggle when the label noise depends on instance features, known as instance-dependent noise, which is more common in real-world datasets.

In this work, we propose a hybrid learning framework that combines self-supervised pre-training and iterative pseudo-label refinement to tackle instance-dependent label noise. By leveraging self-supervised learning, we establish a noise-agnostic feature foundation, followed by iterative refinement to improve label quality. We demonstrate the effectiveness of our method on benchmark datasets, achieving notable improvements over existing approaches, especially in complex noise scenarios.

\section{Related Works}

Learning from noisy labels has become an important area of research due to its significance in real-world scenarios where datasets often contain mislabeled instances. This section highlights major strategies developed to address noisy labels, including loss correction, sample selection, and recent hybrid methods, with a particular emphasis on approaches to instance-dependent label noise.

\textbf{Loss Correction and Small-Loss Selection}: A prominent approach to handling noisy labels involves correcting the loss using estimated label transition matrices. \citep{patrini2017making} introduced the F-correction technique to estimate these transitions, while \citep{goldberger2017training} used an additional softmax layer to model noise more effectively. Another line of research has focused on leveraging the tendency of deep networks to learn clean samples first. \citep{jiang2018mentornet} and \citep{han2018co} use this property to identify and train on small-loss examples, which are likely to be correctly labeled. However, these approaches struggle when noise is instance-dependent, making it challenging to apply uniform selection criteria.

\textbf{Instance-Dependent Noise Handling}: Instance-dependent label noise, where the probability of incorrect labels depends on instance-specific features, is more difficult to handle. Methods like \textit{JoCoR (Joint Co-Regularization)} \citep{wei2020combating} address this by enforcing agreement between two classifiers through a joint regularization term, thereby reducing the effects of instance-dependent noise. \citep{cheng2020learning} introduced a noise transition network to dynamically model feature-dependent noise, providing a more flexible adaptation to label errors.

\textbf{Self-Supervised and Pseudo-Labeling Approaches}: Recent research has leveraged self-supervised pre-training to improve feature extraction without relying on labels, mitigating the effect of noisy annotations. \textit{MoCo} \citep{he2020momentum} and \textit{BYOL} \citep{grill2020bootstrap} are prominent examples, which learn feature representations without the need for annotated data. Additionally, pseudo-label refinement techniques, such as those proposed by \citep{arazo2019unsupervised} and \citep{li2020dividemix}, iteratively replace noisy labels with refined predictions to enhance model robustness against label noise.

\textbf{Hybrid Approach Motivation}: Our proposed method combines self-supervised pre-training with pseudo-label refinement to effectively tackle instance-dependent noise. Self-supervised learning aids in learning robust, noise-agnostic features, while iterative pseudo-label refinement helps in progressively improving label quality. This hybrid approach aims to address the limitations of existing methods and enhance model performance in the presence of complex, instance-dependent noise.

\section{Methodology}

In this section, we describe our proposed hybrid training pipeline designed to mitigate the effects of instance-dependent noise (IDN) in datasets. Our approach integrates self-supervised learning (SSL) using SimCLR with iterative pseudo-label refinement, creating a robust framework to handle noisy labels through a combination of feature pretraining, adaptive pseudo-label selection, and dynamic data augmentation.

\subsection{Pipeline Overview}

\subsubsection{SimCLR-based Self-Supervised Pretraining}

We begin by leveraging SimCLR-based contrastive learning \citep{chen2020simclr} to develop noise-resilient feature representations. This pretraining phase aims to learn general-purpose features that are robust to noise, independent of labeled data.

\paragraph{Contrastive Learning Objective}

SimCLR employs contrastive learning to maximize agreement between differently augmented views of the same image while minimizing agreement between views of different images in the batch. By treating augmented views of the same image as positive pairs and different images as negative pairs, SimCLR learns feature embeddings that are invariant to typical transformations.

\begin{figure*}[t]
    \centering
    \includegraphics[scale = 0.5]{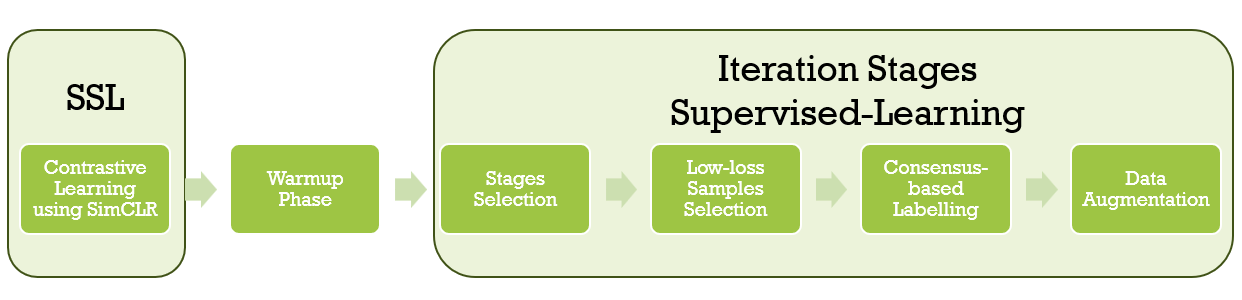}
    \caption{Model Pipeline Flowchart}
\end{figure*}

\paragraph{Augmentation Strategy}

To generate diverse positive pairs, we apply strong augmentations, including \texttt{RandomResizedCrop}, \texttt{ColorJitter}, \texttt{GaussianBlur}, and \texttt{RandomGrayscale}. These augmentations help the model learn features that are invariant to input variations, which is essential for robust feature learning in noisy environments.

\paragraph{Training Configuration}

Pretraining is conducted for 30 epochs. The resulting embeddings are then used as the initial feature representations for downstream supervised learning, establishing a foundation that is inherently resistant to noise.

\subsubsection{Warmup Phase}

Following pretraining, the model undergoes a \textit{warmup phase} for 5 epochs. During this phase, the model is trained on the noisy dataset using cross-entropy loss.

\paragraph{Purpose of Warmup}

The objective of the warmup phase is to initialize the model for supervised learning by capturing basic patterns from the data without overfitting to noisy labels. The warmup duration of 5 epochs is empirically chosen to ensure adequate feature learning while avoiding significant overfitting.

\subsection{Iterative Training with Stages and Pseudo-Label Refinement}

The central component of our methodology is an iterative pseudo-label refinement mechanism. This process is conducted over multiple iterations, each containing predefined \textit{stages} for pseudo-label selection and refinement.

\subsubsection{Stage Definition}

The training process is divided into specific \textit{stages}, defined as sets of epochs within each iteration.

\paragraph{Stage Concept}

In the first iteration, epochs 2, 3, and 4 are designated as stages, whereas subsequent iterations utilize epochs 2, 5, and 7. The choice of stages introduces variability in the learning process, balancing consistent monitoring of sample performance (with consecutive epochs) with broader inspection intervals (using more widely spaced epochs).

\paragraph{Motivation for Stage Choice}

The combination of consecutive epochs (e.g., 2, 3, 4) and more spaced epochs (e.g., 2, 5, 7) ensures a dynamic learning process. Consecutive epochs help identify consistent sample performance over shorter periods, while more spaced epochs allow for an assessment of evolving sample behavior over time, enhancing the robustness of sample selection.

\subsubsection{Loss Threshold Filtering}

During each stage, the model calculates the cross-entropy loss for all training samples.

\paragraph{Sample Selection}

Samples with a loss below a predefined threshold (set to 1 in our experiments) are considered confidently predicted and are selected for further analysis. This threshold was determined empirically to balance the trade-off between selecting clean samples and retaining a sufficient amount of training data for generalization.

\subsubsection{Pseudo-Label Generation}

At the end of each iteration, samples selected from each stage are analyzed to determine their suitability for pseudo-labeling.

\paragraph{Consensus-Based Labeling}

Only samples that are consistently selected across all stages within an iteration are assigned pseudo-labels. This consensus-based labeling strategy ensures that only stable, confidently predicted samples are used for pseudo-labeling, thereby reducing the risk of propagating erroneous labels.

\paragraph{Retention of Original Labels}

Samples that do not meet the consensus criteria retain their original labels. This approach prevents the amplification of label noise by avoiding unreliable pseudo-labeling for uncertain data points.

\subsection{Dynamic Data Augmentation and Training Continuation}

\subsubsection{Augmented Pseudo-Labeling}

After pseudo-labels are generated, we apply data augmentation to enhance the diversity of the training set.

\paragraph{Augmentation Strategy}

The same set of augmentations used during SimCLR pretraining (\texttt{RandomResizedCrop}, \texttt{ColorJitter}, \texttt{GaussianBlur}, \texttt{RandomGrayscale}) is applied to pseudo-labeled samples. This consistent augmentation strategy ensures that the model continues to see diverse transformations, thereby enhancing generalization capability.

\paragraph{Training on Augmented Data}

Augmented versions of pseudo-labeled samples are added back to the training set. This helps the model to learn cleaner data patterns while mitigating overfitting to potentially noisy labels in the original dataset.

\subsubsection{Iterative Training Continuation}

The training set, now enriched with augmented pseudo-labeled data, is used for the subsequent iteration of training. The iterative process continues for a total of four iterations, each involving pseudo-label refinement and retraining. This gradual refinement helps the model improve its understanding of clean data while increasing robustness against noisy labels.

\subsection{Key Innovations of the Proposed Pipeline}

\begin{itemize}
    \item \textbf{SimCLR-based Pretraining for Noise Robustness}: Leveraging contrastive learning during pretraining helps the model learn noise-agnostic features, reducing the impact of label noise in downstream tasks.
    \item \textbf{Stage-Wise Iterative Refinement}: Dividing training into stages helps identify consistent samples over time, leading to more reliable pseudo-labeling and reducing the propagation of erroneous labels.
    \item \textbf{Handling Instance-Dependent Noise}: Unlike approaches that assume independent label noise, our method adapts to IDN through iterative refinement, which dynamically distinguishes between clean and noisy labels.
    \item \textbf{Augmented Pseudo-Labeling}: Combining pseudo-labeled samples with augmentations strengthens the model's ability to generalize clean data patterns, reducing the risk of overfitting to noisy information.
\end{itemize}

\section{Results}

We evaluated the performance of our proposed hybrid method against several state-of-the-art approaches under varying levels of instance-dependent noise (IDN) on the CIFAR-10 and CIFAR-100 datasets. The comparative results are summarized in Table~\ref{tab:results}.

\begin{table}[H]
    \centering
    \caption{Comparison of Our Method with Previous Methods (Accuracy \%)}
    \label{tab:results}
    \begin{tabular}{|c|c|c|c|c|}
        \hline
        \multirow{2}{*}{Method} & \multicolumn{2}{c|}{CIFAR-10} & \multicolumn{2}{c|}{CIFAR-100} \\ 
        \cline{2-5}  
        & 20\% Noise & 50\% Noise & 20\% Noise & 50\% Noise \\ \hline
        Co-teaching \citep{han2018co} & 80.96 & 59.56 & 63.58 & 42.74 \\ \hline
        MentorNet \citep{jiang2018mentornet} & 81.03 & 59.68 & 63.51 & 42.86 \\ \hline
        JoCoR \citep{wei2020combating} & 83.95 & 60.42 & 64.42 & 45.30 \\ \hline
        DivideMix \citep{li2020dividemix} & 90.00 & 75.00 & 77.00 & 57.00 \\ \hline
        \textbf{Our Method} & \textbf{88.41} & \textbf{60.10} & \textbf{63.42} & \textbf{48.57} \\ \hline
    \end{tabular}
\end{table}

\subsection{Performance Overview}

Table~\ref{tab:results} presents the classification accuracy of various methods under two noise levels: 20\% and 50\% IDN on both CIFAR-10 and CIFAR-100 datasets. Our method demonstrates competitive performance, especially notable in the CIFAR-100 dataset under high noise conditions.

\subsection{Comparison with DivideMix}

To further analyze the efficacy of our approach, we integrated our pipeline with DivideMix \citep{li2020dividemix} and compared the combined performance against the standalone DivideMix. The results are shown in Table~\ref{tab:comparison}.

\begin{table}[H]
    \centering
    \caption{Comparison between DivideMix and Our Method when Built on Top of DivideMix (Accuracy \%)}
    \label{tab:comparison}
    \begin{tabular}{|c|c|c|c|c|}
        \hline
        \multirow{2}{*}{Method} & \multicolumn{2}{c|}{CIFAR-10} & \multicolumn{2}{c|}{CIFAR-100} \\ 
        \cline{2-5}  
        & 20\% Noise & 50\% Noise & 20\% Noise & 50\% Noise \\ \hline
        Our Method & 88.41 & 60.10 & 63.42 & 48.57 \\ \hline
        Our Method (with DivideMix) & \textbf{93.02} & 63.35 & 74.17 & 49.51 \\ \hline
    \end{tabular}
\end{table}

Additionally, Table~\ref{tab:dividemix} highlights the baseline performance of DivideMix under the same noise conditions based on results obtained through our implementation, enabling a direct comparison with our proposed approach.

\begin{table}[H]
    \centering
    \caption{DivideMix Performance (Accuracy \%)}
    \label{tab:dividemix}
    \begin{tabular}{|c|c|c|c|c|}
        \hline
        \multirow{2}{*}{Method} & \multicolumn{2}{c|}{CIFAR-10} & \multicolumn{2}{c|}{CIFAR-100} \\ 
        \cline{2-5}  
        & 20\% Noise & 50\% Noise & 20\% Noise & 50\% Noise \\ \hline
        DivideMix \citep{li2020dividemix} & 90.00 & 75.00 & 77.00 & 57.00 \\ \hline
    \end{tabular}
\end{table}

\section{Discussion}

The experimental results elucidate several key insights into the performance and effectiveness of our proposed hybrid method in mitigating instance-dependent label noise.

\subsection{Comparison with Existing Methods}

Our method exhibits robust performance across both CIFAR-10 and CIFAR-100 datasets, particularly under higher noise levels. While DivideMix outperforms our standalone approach in most scenarios, the integration of our method with DivideMix (as shown in Table~\ref{tab:comparison}) leads to superior performance, especially in CIFAR-10 with 20\% noise, achieving an accuracy of 93.02\%. This suggests that our hybrid pipeline can complement and enhance existing noise-robust methods.

\subsection{Effectiveness in High-Noise Scenarios}

Under the most challenging condition of 50\% IDN noise on CIFAR-100, our method achieves an accuracy of 48.57\%, which, although lower than DivideMix's 57.00\%, still represents a significant improvement over other baseline methods. This indicates that while our approach is effective, there is room for further optimization, particularly in handling highly noisy, complex datasets.

\subsection{Integration with DivideMix}

The enhanced performance observed when our method is integrated with DivideMix underscores the complementary strengths of self-supervised pre-training and pseudo-label refinement. The combination leverages DivideMix's robust noise modeling with our noise-agnostic feature learning, resulting in improved classification accuracy. This synergy highlights the potential for our framework to augment existing techniques, fostering more resilient models against label noise.

\subsection{Impact of Self-Supervised Pre-training}

Integrating SimCLR-based self-supervised learning proves instrumental in establishing a strong feature foundation that is less susceptible to label noise. This pre-training phase allows the model to learn meaningful representations without the influence of noisy labels, thereby enhancing the overall robustness during supervised fine-tuning. The effectiveness of SimCLR in our pipeline is evidenced by the competitive performance gains, particularly when combined with other noise-robust methods.

\subsection{Limitations and Future Improvements}

While our method demonstrates competitive performance, especially when integrated with DivideMix, it does not consistently outperform all existing methods across all noise levels and datasets. Specifically, in scenarios with lower noise or less complex datasets, methods like DivideMix alone may offer superior results. Future work could explore adaptive mechanisms to dynamically adjust the balance between self-supervised learning and pseudo-label refinement based on the dataset's noise characteristics.

Additionally, extending our approach to incorporate more advanced self-supervised techniques or leveraging ensemble strategies could further enhance performance. Exploring alternative pseudo-labeling strategies that better capture instance-dependent nuances may also yield improvements.

\subsection{Practical Implications}

The ability to effectively train models on noisy datasets is paramount for real-world applications where perfect labeling is often unattainable. Our hybrid method offers a viable solution by combining robust feature learning with iterative label refinement, thereby enabling the deployment of more reliable models in practical settings. This is particularly relevant in domains such as medical imaging or large-scale data collection, where label noise is prevalent.

\subsection{Future Research Directions}

Building on the current findings, future research could investigate the following avenues:
\begin{itemize}
    \item \textbf{Adaptive Stage Selection}: Developing mechanisms to dynamically select and adjust training stages based on real-time performance metrics.
    \item \textbf{Advanced Self-Supervised Techniques}: Integrating newer self-supervised learning frameworks that may offer superior feature representations.
    \item \textbf{Semi-Supervised Extensions}: Combining our approach with semi-supervised learning to further leverage unlabeled data alongside noisy labels.
    \item \textbf{Domain-Specific Applications}: Applying and tailoring the hybrid method to specialized domains such as natural language processing or medical diagnostics to assess its versatility.
\end{itemize}

\section{Conclusion}

This study addresses the critical challenge of training deep neural networks on datasets with instance-dependent label noise. By integrating self-supervised pre-training with iterative pseudo-label refinement, our proposed hybrid method establishes a robust framework that significantly mitigates the adverse effects of noisy labels. The empirical evaluations on CIFAR-10 and CIFAR-100 datasets demonstrate the effectiveness of our approach, especially when combined with existing methods like DivideMix. While our method shows promising results, particularly in high-noise scenarios, further enhancements and integrations are necessary to achieve state-of-the-art performance across diverse datasets and noise conditions. Future work will focus on refining the pipeline, exploring advanced self-supervised techniques, and extending the approach to a broader range of applications.







\end{document}